\documentclass{article}
\usepackage{ijcai25}

\usepackage{times}
\usepackage{soul}
\usepackage{url}
\usepackage[hidelinks]{hyperref}
\usepackage[utf8]{inputenc}
\usepackage[small]{caption}
\usepackage{graphicx}
\usepackage{amsmath}
\usepackage{amsthm}
\usepackage{amssymb}
\usepackage{booktabs}
\usepackage{algorithm}
\usepackage{algorithmic}
\usepackage[switch]{lineno}
\usepackage[table ]{ xcolor}
\usepackage{array}
\usepackage{listings}
\usepackage{stfloats}
\title{SpatialLLM: From Multi-modality Data to Urban Spatial Intelligence}

\author{
Jiabin Chen$^{1,*}$\and
Haiping Wang$^{1,*}$\and
Jinpeng Li$^{1}$\and
Yuan Liu$^{2,\dagger}$\and
Zhen Dong$^{1,\dagger}$\and
Bisheng Yang$^1$
\affiliations
$^1$Wuhan University 
$^2$Hong Kong University of Science and Technology
\emails
\{chenjb67, hpwang, lijp57, dongzhenwhu, bshyang\}@whu.edu.cn,\
yuanly@ust.hk
\emails
$^{*}$:Equal contribution. $^{\dagger}$:Corresponding authors.
}
\begin{document}

% \maketitle
\twocolumn[{
\renewcommand\twocolumn[1][]{#1}
\maketitle
\begin{center}
    \captionsetup{type=figure}
    \includegraphics[width=1\textwidth]{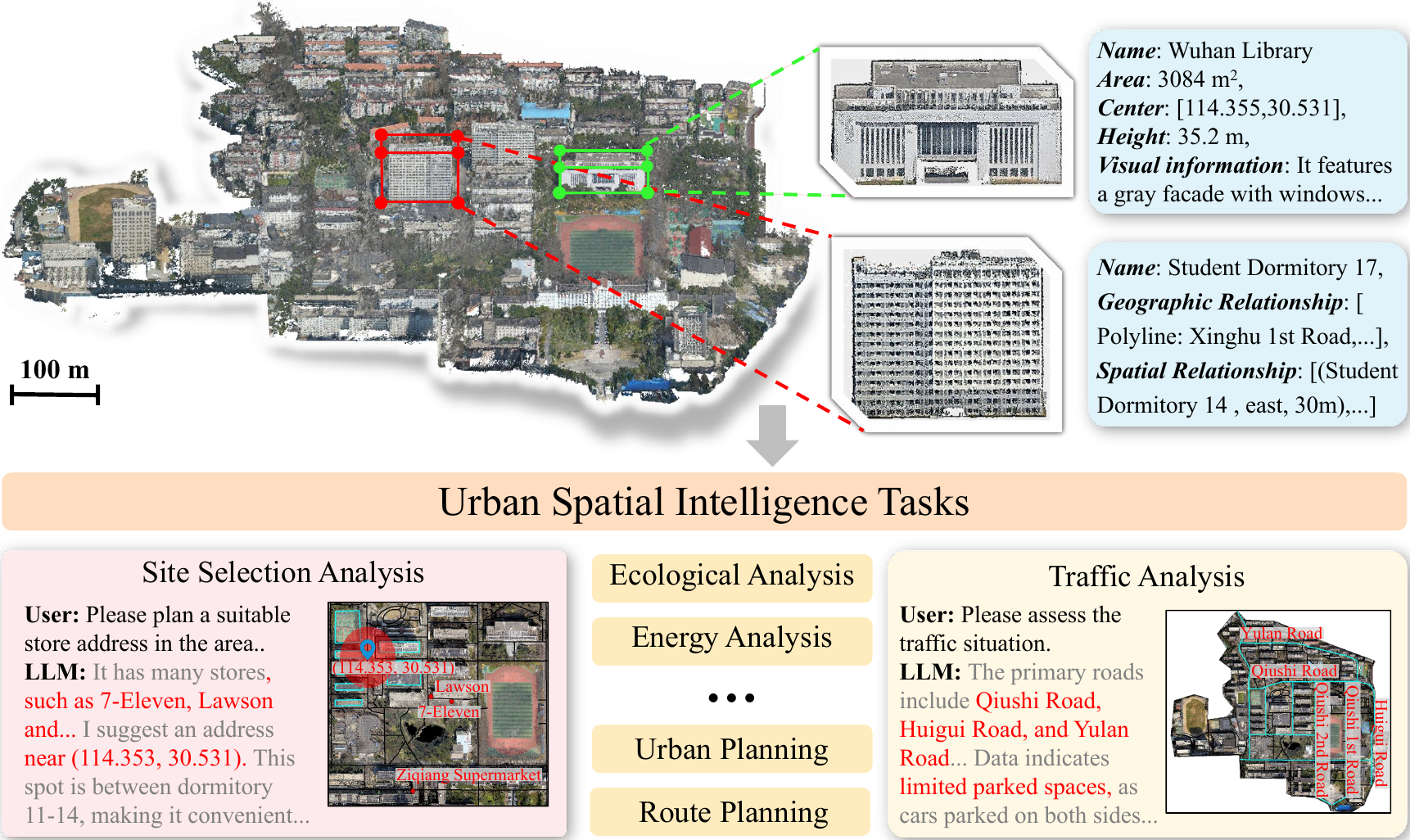}
    \captionof{figure}{We present SpatialLLM, an innovative framework that enables spatial intelligence by transforming multi-modality data into structured text. SpatialLLM first constructs comprehensive scene descriptions from raw spatial data, capturing complex object attributes and spatial relationships within a large outdoor scene. Then, by feeding these structured descriptions into pretrained LLMs, SpatialLLM can perform advanced spatial analysis including site selection, ecological analysis, etc.}
\end{center}
}]
\begin{abstract}
We propose SpatialLLM, a novel approach advancing spatial intelligence tasks in complex urban scenes. Unlike previous methods requiring geographic analysis tools or domain expertise, SpatialLLM is a unified language model directly addressing various spatial intelligence tasks without any training, fine-tuning, or expert intervention.
The core of SpatialLLM lies in constructing detailed and structured scene descriptions from raw spatial data to prompt pre-trained LLMs for scene-based analysis. 
Extensive experiments show that, with our designs, pretrained LLMs can accurately perceive spatial distribution information and enable zero-shot execution of advanced spatial intelligence tasks, including urban planning, ecological analysis, traffic management, etc. 
We argue that multi-field knowledge, context length, and reasoning ability are key factors influencing LLM performances in urban analysis.
We hope that SpatialLLM will provide a novel viable perspective for urban intelligent analysis and management. 
The code and dataset are available at https://github.com/WHU-USI3DV/SpatialLLM.
\end{abstract}

\section{Introduction}
\label{sec1}
%% Labels are used to cross-reference an item using \ref command.
Urban spatial intelligence has long been hindered by the need for significant human knowledge and intervention, constraining its scalability and efficiency~\cite{mai2023opportunities,mai2022towards}. But imagine a paradigm shift—where multi-modality raw spatial data can be seamlessly transformed into suggestions for urban planning, risk analysis, and ecological evaluation. These would fundamentally redefine how we interpret urban information and act upon urban environments, yielding a novel view of our urban management.

To achieve this, some existing methods proposed to build a model, typically the Multimodal Large Language Model (MLLM), that accepts both scene data and textual queries, and provides responses. By training on many scene-based
Question Answering (QA) data, these models~\cite{hong20233d,fu2024scene,wang2023chat,huang2024chat} can handle perception tasks such as grounding and scene understanding.
Another line of works~\cite{gu2024conceptgraphs,zhang2024tag} require no additional training. They propose that by describing indoor scenes in text, these descriptions can be directly input into the pre-trained LLM for scene-based QA leveraging the rich knowledge priors of LLMs. However, the above methods are only validated on small-scale indoor scenes using basic spatial perception tasks such as grounding or captioning~\cite{hong20233d,gu2024conceptgraphs}.

The outdoor scenes present different challenges altogether than their indoor counterparts. 
First, \textit{train-data scarcity} - considering the diversity of urban data (maps, images, point clouds, etc.) and the high cost of annotation, it is difficult to train an MLLM at scale~\cite{miyanishi2023cityrefer}.
Second, \textit{scene complexity} - outdoor scenes feature various objects and complex relationships compared to indoor scenes, making it harder for LLM to understand and process~\cite{lei2023recent,han2024whu}. 
Third, \textit{task diversity} -  outdoor spatial intelligence tasks extend far beyond simple perception, requiring advanced tasks such as urban planning, ecological analysis, and traffic management~\cite{wang2023towards,meng2023spatial}.

To address the above challenges, we propose SpatialLLM — a novel paradigm \textit{requiring no training or tuning, handling complex 3D urban scenes, and processing advanced spatial intelligence tasks given raw urban data}. 
SpatialLLM first constructs textural scene descriptions based on raw multi-modality data, which are then used as prompts for the LLM to support scene-based QA, yielding the following key innovations.
\begin{enumerate}
    \item \textbf{A Multi-modality Data Joint Description Module}. It can automatically extract and fuse information from raw multi-modality urban spatial data. These form rich urban descriptions capturing multi-dimensional attributes of urban entities, such as geometry, texture, and function, as well as relational information among them.
    \item \textbf{LLM-assisted Urban Task Processing}. SpatialLLM feeds the above urban descriptions into LLM to, for the first time, explore the potential of LLMs in handling various advanced urban tasks. It further identifies key factors that influence the performance, specifically multi-field knowledge, context length, and reasoning ability.
    \item \textbf{LLM spatial perception dataset}. A novel dataset comprising multi-modality spatial data with human-annotated and spatial-related QA cases, designed for evaluating the spatial perception accuracy of LLM-based methods.
\end{enumerate}
\section{Related work}

\subsection{Indoor Spatial Intelligence}
With the foundation of extensive and diverse indoor spatial datasets~\cite{chang2017matterport3d,ramakrishnan2021habitat}, research in indoor scene intelligence has developed rapidly. However, early research was primarily task-specific, including semantic segmentation~\cite{lai2022stratified,fan2021scf,peng2023openscene}, visual grounding~\cite{chen2020scanrefer,zhao20213dvg,yang2024exploiting}, question answering~\cite{azuma2022scanqa,ma2022sqa3d}, etc. While these investigations achieved substantial advances, they remained compartmentalized along independent technical trajectories, lacking the unified intelligence framework~\cite{lei2023recentadvancesmultimodal3d} needed for comprehensive spatial reasoning.

The emergence of large language models (LLMs) has substantially accelerated the development of research on indoor spatial intelligence~\cite{ma2024llms}. 
First, LLMs revolutionized dataset construction: In dataset construction, 3D-VisTA~\cite{zhu20233d} leveraged GPT-3 to generate over 270,000 scene descriptions, effectively overcoming the limitations of manual annotation, while 3D-LLM~\cite{hong20233d} utilized LLMs to construct a comprehensive dataset that includes multiple tasks. Scene-LLM~\cite{fu2024scene} created an extensive dataset, comprising 190,000 3D-visual-language pairs from egocentric viewpoints and approximately 500,000 pairs of scene-level data. Furthermore, EmbodiedScan~\cite{wang2024embodiedscan} and MMScan~\cite{lyu2024mmscan} expanded annotation scales to millions of entries, covering multi-level semantic information from object recognition and spatial relationships to functional reasoning. 
Second, LLMs enabled unified multimodal understanding architectures that transcend task boundaries. 3D-LLM~\cite{hong20233d} pioneered the integration of 3D point cloud features into LLM, establishing a unified processing framework for a variety of 3D-related tasks. Scene-LLM~\cite{fu2024scene} uniquely combines scene-level and egocentric 3D information to enable effective reasoning and interaction in indoor scenes. Chat-3D~\cite{wang2023chat} and LL3DA~\cite{chen2024ll3da} introduced object-centric interaction mechanisms, enabling models to process diverse queries based on specific objects. 
Recent work has enhanced indoor spatial intelligence through knowledge augmentation~\cite{chandhok2024scenegptlanguagemodel3d}, perceptual reasoning~\cite{Chen_2024_CVPR}, multimodal data integration~\cite{huang2023embodied}, among others.

However, these approaches remain predominantly focused on  indoor scenes, leaving a significant gap in addressing the more complex challenges of outdoor urban spatial intelligence.

\begin{figure*}[!ht]
    \centering
    \includegraphics[width=1\textwidth]{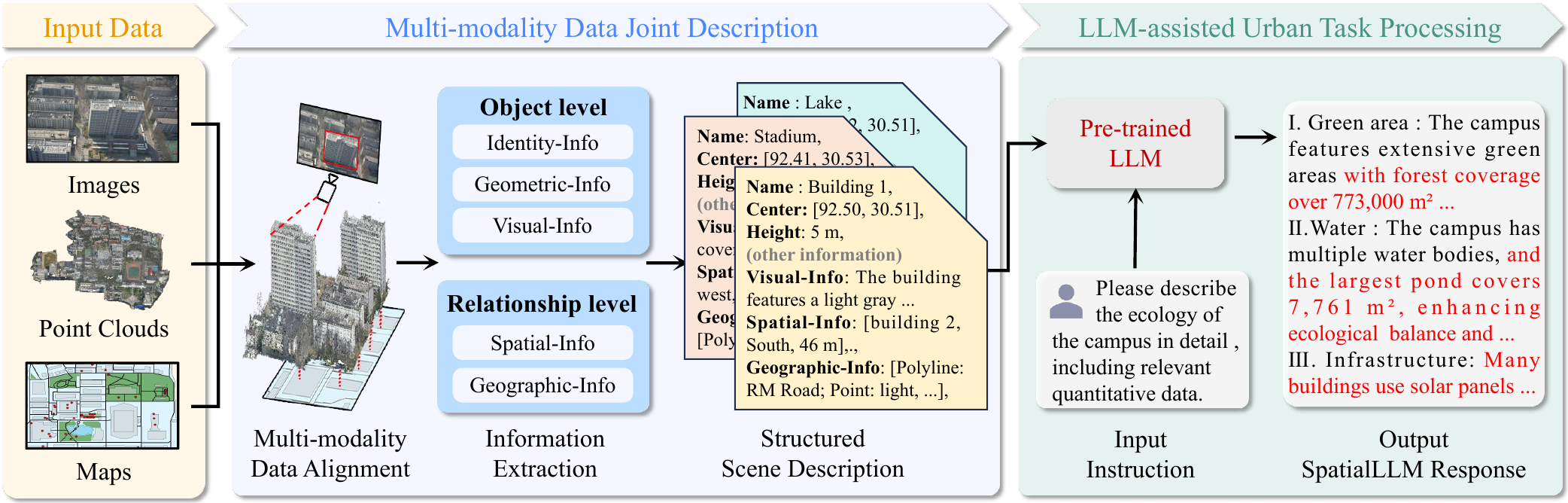}
    \caption{The overview of SpatialLLM. SpatialLLM conducts advanced spatial intelligence tasks with raw urban data inputs.}
    \label{fig:pipeline}
\end{figure*}

\subsection{Urban Spatial intelligence}
Research on urban spatial intelligence has progressed more gradually compared to indoor research, constrained by environmental complexity and large-scale spatial modeling challenges. Although abundant urban-scale 3D datasets~\cite{yang2023urbanbis,han2024whu} have been established, datasets for downstream urban spatial intelligence tasks remain limited, primarily due to prohibitive annotation costs and technical challenges. 
Despite these constraints, notable advancements~\cite{bie2025hyperg,du2025robust,yu2024coastal} have emerged in specific domains. CityRefer~\cite{miyanishi2023cityrefer} annotated the SensatUrban dataset to train a baseline for city-scale 3D visual grounding, while CityAnchor~\cite{licityanchor} significantly enhanced performance through an innovative two-stage MLLM architecture. Sun et al.~\cite{sun20243d} introduced the City-3DQA dataset and SG-CityU model for urban-scale 3D question answering. Furthermore, Jie et al.~\cite{feng2024citygpt} developed CityInstruction, which incorporates urban knowledge to augment LLMs' spatial comprehension for city-related tasks.

While the above approaches focus primarily on processing visual scene data, a parallel line of research has explored the potential of LLMs for urban geospatial intelligence. Geographic information (e.g., landmarks, road networks) plays a crucial role in urban spatial~\cite{su2024multimodal,huang2024crowdsourced}. Recent research has extended LLMs for urban geospatial intelligence~\cite{li2023autonomous}, following two main approaches. The first approach integrates LLMs with GIS tools for zero-shot urban geospatial analysis~\cite{zhang2023geogpt}, enabling natural language interactions with geographic databases and supporting vector-based spatial operations for complex geographic queries. 
The second approach trains specialized models on large-scale urban spatial datasets~\cite{feng2024citygpt,zhang2024bb,yan2024georeasoner}, showing promising results in tasks such as location recommendation. However, these methods are limited to geographic information and do not effectively incorporate real-world scene data.

In summary, urban spatial intelligence research confronts three significant limitations. First, the scarcity of training data constrains model generalization capabilities across complex outdoor environments. Second, existing methods have not effectively bridged the gap between visual scene understanding and geospatial analysis, limiting their ability to achieve greater urban spatial intelligence. Third, current research predominantly adheres to single-task paradigms or low-level tasks, lacking a comprehensive framework capable of unifying multiple advanced intelligence tasks.

To address these challenges, we propose SpatialLLM, a novel framework for urban spatial intelligence without additional training, fine-tuning or expert intervention. By transforming multi-modality urban data into structured scene descriptions to prompt pre-trained LLMs, SpatialLLM not only handles spatial perception tasks, but also enables zero-shot execution of advanced spatial intelligence tasks.

\section{SpatialLLM Overview}

As shown in Fig.~\ref{fig:pipeline}, given raw spatial data—including maps, images, and point clouds—SpatialLLM first employs a Multi-modality Data Joint Description (MDJD) module in Sec.~\ref{sec:mdjd} to extract scene information. Specifically, MDJD captures identity information from map data, geometric information from point cloud data, and visual information from image data. These, along with relationship information between objects, are integrated into a textual Structured Scene Description (SSD). Then, in LLM-assisted Urban Task Processing (LUTP) of Sec.\ref{sec:lutp}, we resort to the knowledge of pre-trained LLMs to perform complex spatial intelligence tasks given SSD, and analyze the key factors influencing the performances.

\section{Multi-modality Data Joint Description (MDJD) Module}
\label{sec:mdjd}

Multi-modality spatial data, including maps, images, and point clouds, provides complementary urban information from multiple dimensions. To construct more comprehensive and reliable scene descriptions, it is essential to widely extract information from various types of modality data.
As shown in Fig.~\ref{fig:pipeline}, MDJD aligns the multi-modality data and automatically extracts rich textural descriptions of urban objects and their relationships as follows.

\subsection{Multi-modality Data Alignment}
\label{sec:pre}

To align point clouds with maps, we project and rasterize the point clouds into a top-view image. Then we solve an affine alignment between the projected image and map with Least Square method on manually selected correspondences. To align images and point clouds, we employ Structure-from-Motion~\cite{schoenberger2016sfm} and Multi-View Stereo~\cite{schoenberger2016mvs} to reconstruct the 3D scene from multi-view images. Then we solve a 7-DOF transformation to align the reconstructed and scanned point clouds same as the above method. 

\subsection{Information Extraction of Multi-modality data}
\label{sec:info}

\subsubsection{4.2.1 Object information}
\label{sec:obj info}
\textbf{Identity information from map data.} Map data such as OpenStreetMap (OSM)~\cite{haklay2008openstreetmap} provides rich human-annotated attribute information about urban objects, including their names, function and type, etc. Since these contents are inherently represented in text, we can directly extract and utilize them as a part of SSD.
However, for complex objects such as buildings, map data alone is insufficient to capture detailed information such as texture, shape, and height. Therefore, it is necessary to incorporate additional information from images and point clouds.

\textbf{Geometric information from point clouds.} Given that the point clouds data has been aligned with the map data, we can directly use polygon boundaries from map data to segment individual objects in the point cloud. This allows us to extract precise 3D geometric information, such as object's center coordinates, height, area, volume, bounding boxes, etc., that cannot be derived from map data alone. 
This detailed geometric information provides essential spatial information that significantly enhances the richness and precision of the scene description, allowing for more accurate spatial reasoning in subsequent tasks.

\begin{figure*}[!ht]
    \centering
    \includegraphics[width=1\linewidth]{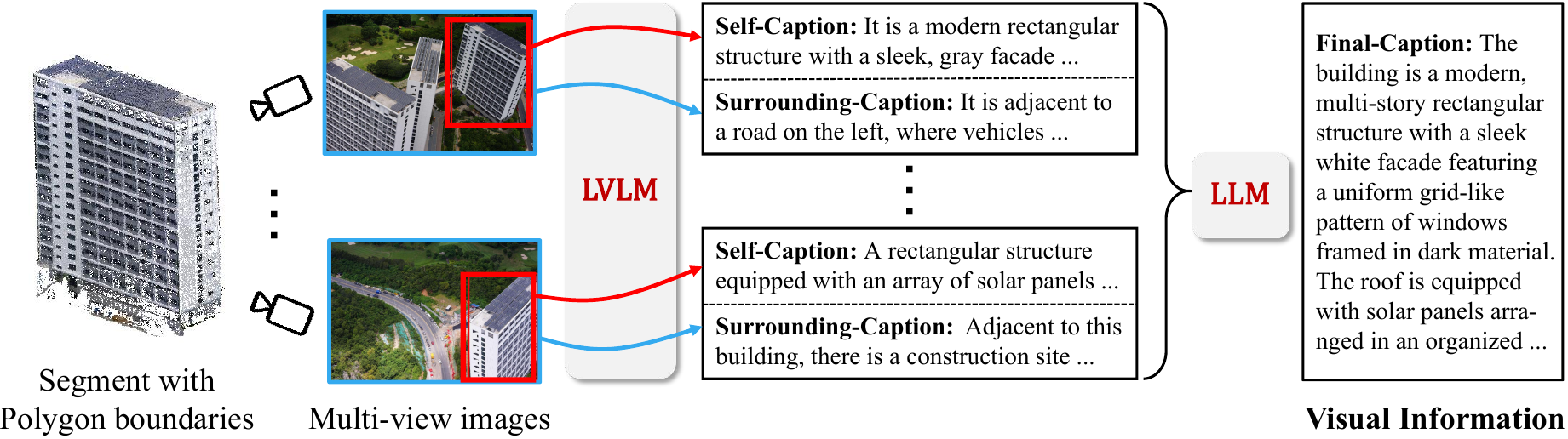}
    \caption{Captioning polygon-type objects using multi-view images.}
    \label{fig:Multi-view}
    % \vspace{-10pt}
\end{figure*}

\textbf{Visual information from images.} Object textures and fine-grained details are difficult to obtain from map and point cloud data. We thus resort to image data for enriched descriptions. Specifically, we segment the target objects within the reconstructed point cloud from images, and then project them onto the multi-view images to generate bounding boxes for the target.
As in Fig.~\ref{fig:Multi-view}, for each image that captures the target, we first use a Large Vision-Language Model (LVLM) to broadly describe the surrounding regions of the target object, called Surrounding-Caption. Then, we isolate the target object and employ LVLM to generate a detailed, object-specific description, denoted as Self-Caption.
Finally, we integrate the Self- and Surrounding-Captions from multi-view images into an LLM to summarize a comprehensive and compact textural visual information for the target.

\subsubsection{4.2.2 Relationship Information}
In addition to describing individual objects, we further describe the relationships between objects, especially between neighboring objects. This relationship description can be categorized into two aspects: spatial relationship information in point clouds and geographic topological relationship information in maps.

\textbf{Spatial relationship}. For each object $q$, we first retrieve its neighboring objects $\mathcal{N}$ within point clouds and summarize their relative information as $\{( N_i, \theta_i, \delta d_i) | \forall{o_i}\in \mathcal{N}\}$, where $N_i$ means the name of its neighboring object $o_i$, $\theta_i$ represents the direction of $o_i$ relative to $q$,  $\delta d_i$ denotes the distance between $q$ and $o_i$.

\textbf{Geographic topology relationship}. We create a buffer around the query object's outline in map data by a certain distance. We record the information, specifically the name and distance, of the objects that intersect with the expanded area.

\subsection{Structured Scene Description}
\label{sec:SSD}
We organize the aforementioned object and relationship information into textual structured object descriptions. While for tiny objects (e.g., bus stops, traffic signals) and polyline-type objects (e.g., vehicular roads, sidewalks), we find that it is hard to recognize them from images or point clouds due to their slim shape and small-scale. Thus, we primarily use their annotation information from the map data, specifically the name and coordinates.
Subsequently, all objects are organized and recorded according to their object IDs to form the final Structured Scene Description (SSD), which provides a comprehensive, accurate, and structured description of the urban scene.

\section{LLM-assisted Urban Task Processing}
\label{sec:lutp}

In this section, the SSD is fed into pre-trained Large Language Models (LLMs) to provide urban contextual information. We aim to evaluate the LLMs’ ability in understanding urban spatial contexts and performing complex urban intelligence tasks by carefully designed QA tasks. Specifically, Sec.~\ref{sec:Ep} introduces a novel dataset qualifying spatial awareness of LLMs. Sec.~\ref{sec:base} examines whether the SSD can effectively provide accurate spatial distribution information to LLMs. 
Sec.~\ref{sec:llm} investigates the factors that influence the LLMs’ fundamental spatial awareness.
Sec.~\ref{sec:adv} evaluates whether SpatialLLM can execute advanced spatial intelligence tasks such as path planning and scene analysis.

\subsection{LLM spatial perception dataset}
\label{sec:Ep}

\begin{figure*}[]
    \centering
    \includegraphics[width=0.9\textwidth]{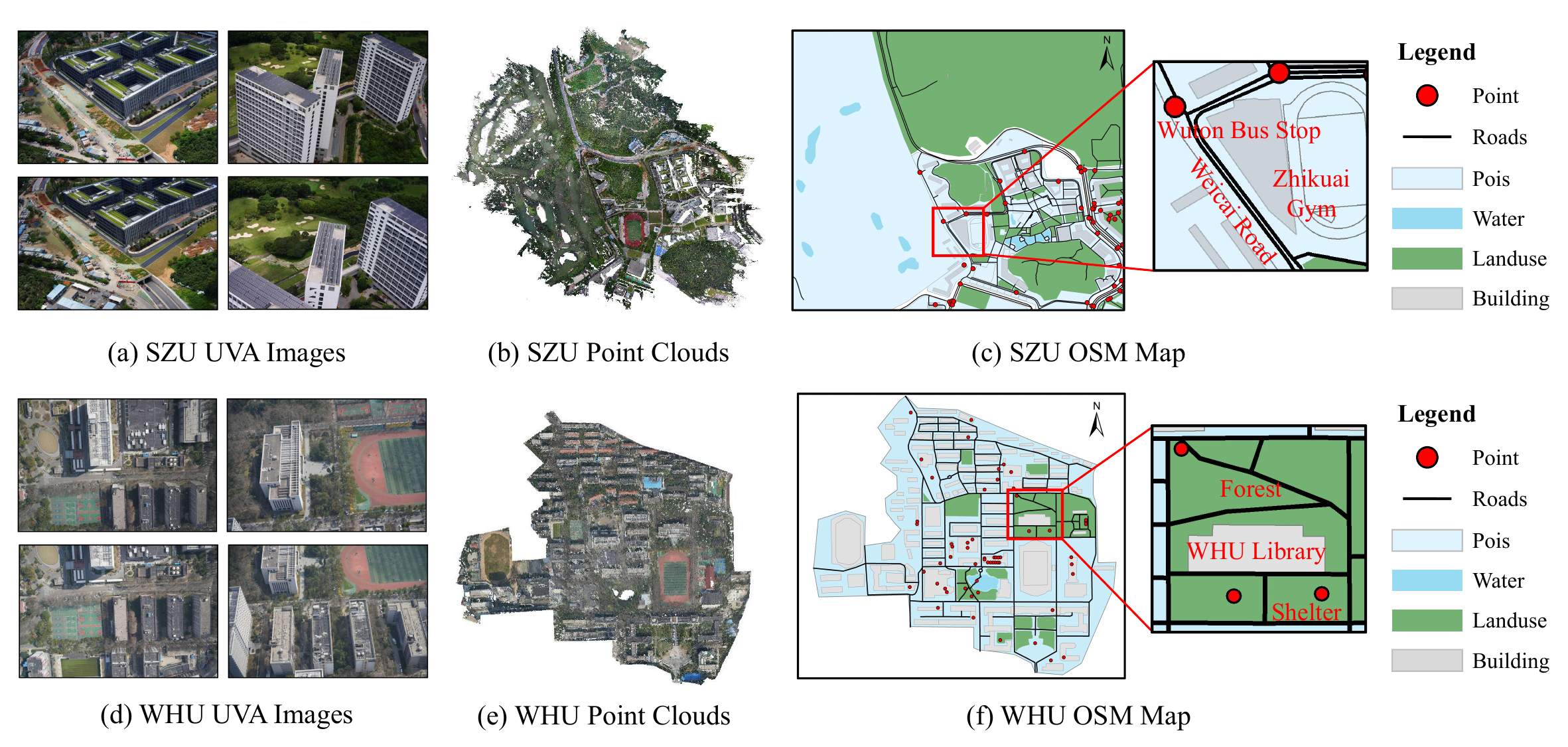}
    \caption{The overview of the raw data of two evaluation scenes.}
    \label{fig:data}
\end{figure*}

We annotated a novel dataset to evaluate the spatial awareness of LLMs. 

\textbf{Data acquisition.} We employ dense UAV imagery data acquired through oblique photography as image data, OpenStreetMap (OSM)~\cite{haklay2008openstreetmap} as map data, and point clouds from UAV reconstruction. We select two evaluation scenes, the Lihu Campus of Shen Zhen University (SZU)~\cite{yang2023urbanbis} covering a 1.3km$^2$ area and the Information Faculty of Wuhan University (WHU) covering a 0.6km$^2$ area, as our evaluation areas. These two sites represent distinct urban environments—SZU is characterized by sparse buildings, abundant vegetation, and a complex road network, while WHU features dense buildings and a more regularized road network. 
Fig.~\ref{fig:data} provides an overview of them.

\textbf{Task design}. We designed five categories to qualify the spatial perception accuracy of LLMs.

\begin{itemize}
    \setlength{\itemsep}{0pt}
    \setlength{\parsep}{0pt}
    \setlength{\parskip}{0pt}
    \item \textit{Distance perception:} Tasks that evaluate the ability to calculate or estimate spatial distances between entities, e.g. ``Calculate the straight-line distance from A to B. ''.
    \item \textit{Directional awareness:} Tasks that assess understanding of relative orientations and cardinal directions, e.g. ``If you are at A, which direction would you walk to reach B? ''.
    \item \textit{POI area recognition:} Tasks that evaluate the ability to identify relevant locations based on user-provided coordinates of interest, e.g. ``Which building is closest to the coordinates (114.30, 30.50)? ''.
    \item \textit{Path selection:} Tasks that evaluate route perception and topological understanding, e.g. ``Which road is the shortest from A to B? ''.
    \item \textit{Grounding:} Tasks that evaluate the ability to locate specific targets based on user-provided descriptive features, e.g. ``Which building has a white roof and black chimneys? ''.
\end{itemize}

Specifically, for each task category, we manually constructed 20 questions to build a spatial perception QA evaluation dataset. Each question includes four options, with only one correct answer. The LLM needs to select an answer based on the constructed structured scene description.

\textbf{Metric design}. For each question, we implemented a standardized output format of ``Option\#Reasoning'' to ask LLM to choose one option among the four candidate answers. Furthermore, to ensure a reliable assessment, we allow a ``F\#Reasoning'' answer when LLM finds none of the four options is right, rather than making arbitrary selections. We report the average correct ratio among the questions.

\subsection{Can LLMs perceive the scene with SSD prompt?}
\label{sec:base}

\textbf{Baselines}. Besides SSD, we also prompt an LLM with scene descriptions from two baselines: the OSM description using only OpenStreetMap data, while the ConceptGraph~\cite{gu2024conceptgraphs} method constructs scene descriptions using only imagery data. 

\textbf{Results}. The experimental results in Table~\ref{tab:baseline} show that LLM, specifically Claude-3.5-Sonnet~\cite{TheC3}, achieves an overall accuracy of 0.74 and 0.79 with SSD on the two datasets, largely exceeding 0.25, confirming its capacity for meaningful spatial information interpretation rather than randomly selecting an answer among four candidate choices. Moreover, SSD prompt yields significantly better accuracy than the OSM (0.54/0.58) and ConceptGraph (0.35/0.59) scene prompt, which confirms the description efficiency of our designs.
Moreover, the ablation studies also confirm that both multi-modality object information and object relational information provide LLMs with more accurate scene perception capabilities. We observe that removing identity information leads to a large performance drop. This is because many queries typically mention object names, which can only be contained in identity information.

\begin{table*}[!ht]
    \centering
    \resizebox{1\linewidth}{!}{%
    \begin{tabular}{l|cccccc}
        \toprule
          Scene Description & Distance&  Directional&  POI &  Path&  Grounding&  Overall\\
         \midrule
         OSM & 0.45 / 0.50 & 0.75 / \textbf{0.90 }& \textbf{0.85} / 0.65& 0.45 / 0.70& 0.20 / 0.15& 0.54 / 0.58 \\
         
         ConceptGraph & 0.20 / 0.45& 0.50 / 0.55& 0.40 / 0.75& 0.20 / 0.40& 0.45 / 0.80& 0.35 / 0.59\\
                  
          Ours (SSD) & \textbf{0.75} / \textbf{0.65} & \textbf{0.85} / 0.85& \textbf{0.85} / \textbf{0.80}& \textbf{0.50} / \textbf{0.80}& \textbf{0.75 }/ \textbf{0.85}& \textbf{0.74 / 0.79}\\
         \midrule
          Ours w/o Identity-Info & 0.05 / 0.15 & 0.15 / 0.15 & 0.45 / 0.40& 0.20 / 0.05& 0.30 / 0.15& 0.23 / 0.18\\
          Ours w/o Geometric-Info & 0.45 / 0.45 & 0.20 / 0.45&0.50 / 0.45 &0.30 / 0.60 &0.65 / 0.80 &0.42 / 0.55\\
          Ours w/o Visual-Info & 0.80 / 0.75 & 0.85 / 0.85& 0.90 / 0.85& 0.55 / 0.70& 0.20 / 0.10& 0.66 / 0.65\\
          Ours w/o Relationship-info & 0.60 / 0.60 &0.90 / 0.80& 0.85 / 0.90& 0.45 / 0.65& 0.80 / 0.75& 0.72 / 0.74\\
         \bottomrule 
        \end{tabular}
        }
    \vspace{-5pt}
    \caption{Spatial perception QA accuracy of SZU/WHU by prompting Claude-3.5-Sonnet with scene descriptions.}
    \vspace{-5pt}
    \label{tab:baseline}
\end{table*}

\subsection{What affects LLM’s spatial analysis ability?}
\label{sec:llm}
To investigate factors influencing LLM’s spatial analysis capabilities, we evaluate various LLMs using our constructed spatial perception QA evaluation dataset, including close-source models such as GPT-4o~\cite{achiam2023gpt}, Claude~\cite{TheC3}, and Qwen~\cite{bai2023qwen}, as well as open-source models, including LLama~\cite{dubey2024llama} and Deepseek~\cite{liu2024deepseek}. With the result shown in Table~\ref{tab:exp}, we observe that all LLMs can perceive scene context from SSD for accuracy much higher than 25\% and further have the following findings.

\begin{table*}
    \centering
    \newcolumntype{G}{>{\color{gray!90}}c}
    \resizebox{\linewidth}{!}{
    \begin{tabular}{c|GGGG|ccccccc}
        \toprule
         Model  & Rsn. & Cont.& Math & MMP. & Distance&  Directional&  POI &  Path&  Grounding&  Overall\\
         \midrule
         Qwen-Max & - & 32K& - &  76.1 & 0.40 / 0.35& 0.55 / 0.20& 0.55 / 0.70 & 0.35 / 0.50&  0.40 / 0.50&  0.45 / 0.45\\
         Qwen-Plus & - & 128K& - &  -  &0.45 / 0.35& 0.80 / 0.75& 0.75 / 0.60 & 0.40 / 0.55& 0.45 / 0.35& 0.57 / 0.52\\
         Llama-3.3-70B & -  &  128K& 77.0 & 68.9  & 0.60 / \textbf{0.70}& 0.55 / 0.60& 0.65 / 0.65& 0.40 / 0.45 & 0.40 / 0.25& 0.52 / 0.53\\
         GPT-4o-mini & - & 128K&70.2  & -  &0.50 / 0.40& 0.25 / 0.30& 0.30 / 0.45& 0.15 / 0.25& 0.35 / 0.35& 0.31 / 0.30\\
         DeepSeek-V3  & - & 128K & 90.2& 75.9  & 0.55 / 0.30& 0.60 / 0.40& 0.80 / 0.75& 0.50 / 0.50& 0.60 / 0.70& 0.61 / 0.53\\
         GPT-4o& - & 128K& 76.6  &  72.6    &0.45 / 0.40 & 0.25 / 0.35& 0.65 / 0.70& 0.40 / 0.50& 0.65 / 0.75& 0.48 / 0.54\\
         Claude-3.5-Sonnet& - & 200K& 78.3  & 78.0  &\textbf{0.75} / 0.65& \textbf{0.85} / \textbf{0.85}& \textbf{0.85} /\textbf{ 0.80} & \textbf{0.50} / \textbf{0.80}& \textbf{0.75} / \textbf{0.85}& \textbf{0.74} / \textbf{0.79}\\     
         \toprule
         Deepseek-R1  & \checkmark & 64K & 97.3 & 84.0  &0.90 / 0.75&  0.70 / 0.80 &  0.85/ 0.70 &  0.40 / 0.70&  0.65 / 0.65&  0.70 / 0.72\\
         GPT-o1 & \checkmark  & 200K & 96.4 & 91.0  &\textbf{0.95} /\textbf{ 0.80}&  \textbf{0.95} / \textbf{0.95} &  0.90/ \textbf{0.85} &  0.50 / \textbf{0.85}&  \textbf{0.80} / \textbf{0.85}&  0.82 / \textbf{0.86}\\
         Claude-3.7-Sonnet  & \checkmark & 200K& 96.2 & - &\textbf{0.95} /\textbf{ 0.80}&  0.85 / 0.85 &  \textbf{1.00}/ 0.80 &  \textbf{0.65} / 0.85&  0.75 / 0.80&  \textbf{0.84} / 0.83\\
         \bottomrule 
    \end{tabular}}
    \vspace{-5pt}
    \caption{Spatial perception QA accuracy of different LLMs on SZU/WHU datasets. Here we also report whether the LLM
    is trained for reasoning (Rsn.), the input context length limitation of each LLM (Cont.), the performance of LLMs on typical Math, MMLU-Pro (MMP) benchmarks for comparison.}
    \vspace{-5pt}
    \label{tab:exp}
\end{table*}

\textbf{Spatial tasks require multi-field knowledge.} Our evaluation results are overall consistent with those of a widely-used benchmark MMLU-Pro~\cite{wang2024mmlu}, which comprehensively evaluates the LLM's abilities with a diverse range of questions including mathematics, computer science, etc. 
This not only validates the effectiveness of our evaluation but also indicates that spatial perception tasks are inherently complex and require LLM to possess multi-field knowledge. Intuitively, even a basic distance perception task requires the model to have mathematical computation skills, as well as fundamental knowledge of latitude and longitude. 
Thus Deepseek excels in mathematics, but its spatial perception accuracy is significantly lower than that of Claude.

\textbf{A larger context window length is necessary}. Scene descriptions are often lengthy. In our evaluation, there are 17.3k/34.1k tokens for SSD of SZU/WHU dataset, requiring LLMs to effectively process long-sequence input and capture long-range information. By comparing Qwen-Max and Qwen-Plus, we found that, under the same model parameter size (72B), Qwen-Plus, which can handle a text length of 128K, achieved a 10\% improvement in spatial perception accuracy. Moreover, the model with a 200K context window generally outperformed the 128K models.

\begin{figure*}[!ht]
    \centering
    \includegraphics[width=0.9\linewidth]{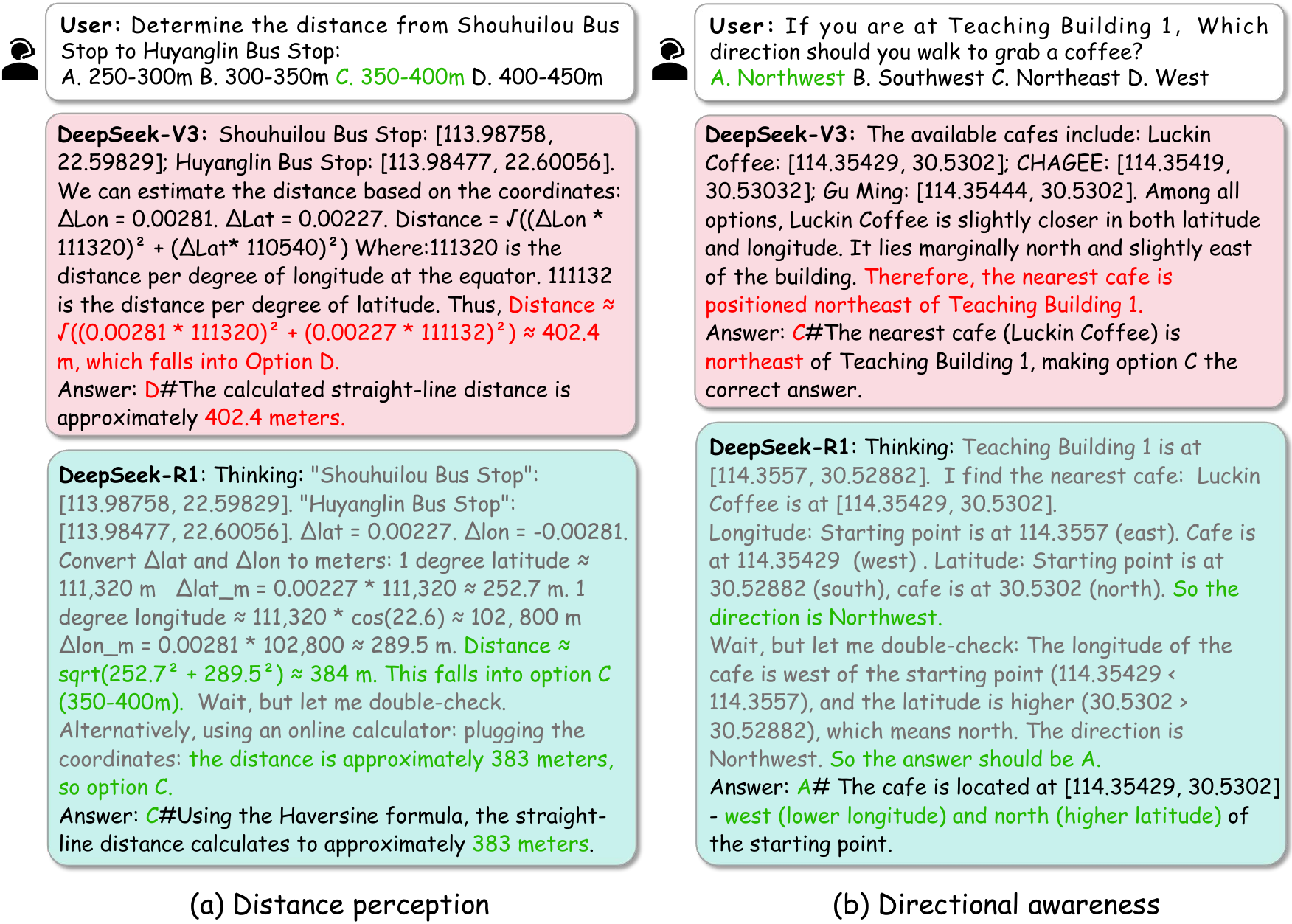}
    \caption{Comparison of spatial reasoning capabilities between the DeepSeek-V3(red) and DeepSeek-R1(green) models. (a) shows a distance perception task where V3 uses simplified Euclidean calculations yielding incorrect results (402.4m), while R1 applies professional Haversine formula with latitude adjustments, correctly calculating 383m. (b) shows that V3 mistakenly identifies the result without position comparison, while R1 correctly determines ``northwest" through systematic longitude-latitude analysis.}
    \label{fig:v3-r1}
\end{figure*}

\textbf{Reasoning capability is crucial}. Table~\ref{tab:exp} clearly shows that models with reasoning abilities can effectively outperform their counterparts. Despite having a shorter context length (64K vs. 128K), DeepSeek-R1 achieved a 14\% accuracy improvement compared to DeepSeek-v3. Furthermore, Claude-3.7 and GPT-o1 achieved an impressive 84\% of the best accuracy. In Fig.~\ref{fig:v3-r1}, we use two examples to validate why reasoning plays an effective role:
\begin{enumerate}
    \item \textbf{Multiple validation mechanism:} DeepSeek-R1 performs ``double-check'' as in Fig~\ref{fig:v3-r1}, which cross-validates the results through multiple computational methods and improves the reliability of the results.
    \item \textbf{Accurate chain-of-thought:} Deepseek-R1 constructs a more accurate ``step-by-step'' CoT reasoning path, yielding fewer errors. In Fig.~\ref{fig:v3-r1}(b), DeepSeek-R1 begins with ``locate the teaching building'' for the following detailed coordinate comparison, but DeepSeek-v3 fails to make such a comparison.
    \item \textbf{Application of expertise:} DeepSeek-R1 effectively leverages a stronger geographic expertise, e.g., choosing the professional Haversine's formula for distance calculation in Fig.~\ref{fig:v3-r1}(a), whereas DeepSeek-V3 uses a simplified Euclidean distance calculation without taking the curvature of the Earth into account.
\end{enumerate}

In summary, enhancing the LLM’s multi-field knowledge, expanding the model's context length, and improving its reasoning abilities can effectively improve the spatial perception accuracy.

\subsection{How well does SpatialLLM handle advanced spatial intelligence tasks?}
\label{sec:adv}

\begin{figure*}[!ht]
    \centering
    \includegraphics[width=0.92\linewidth]{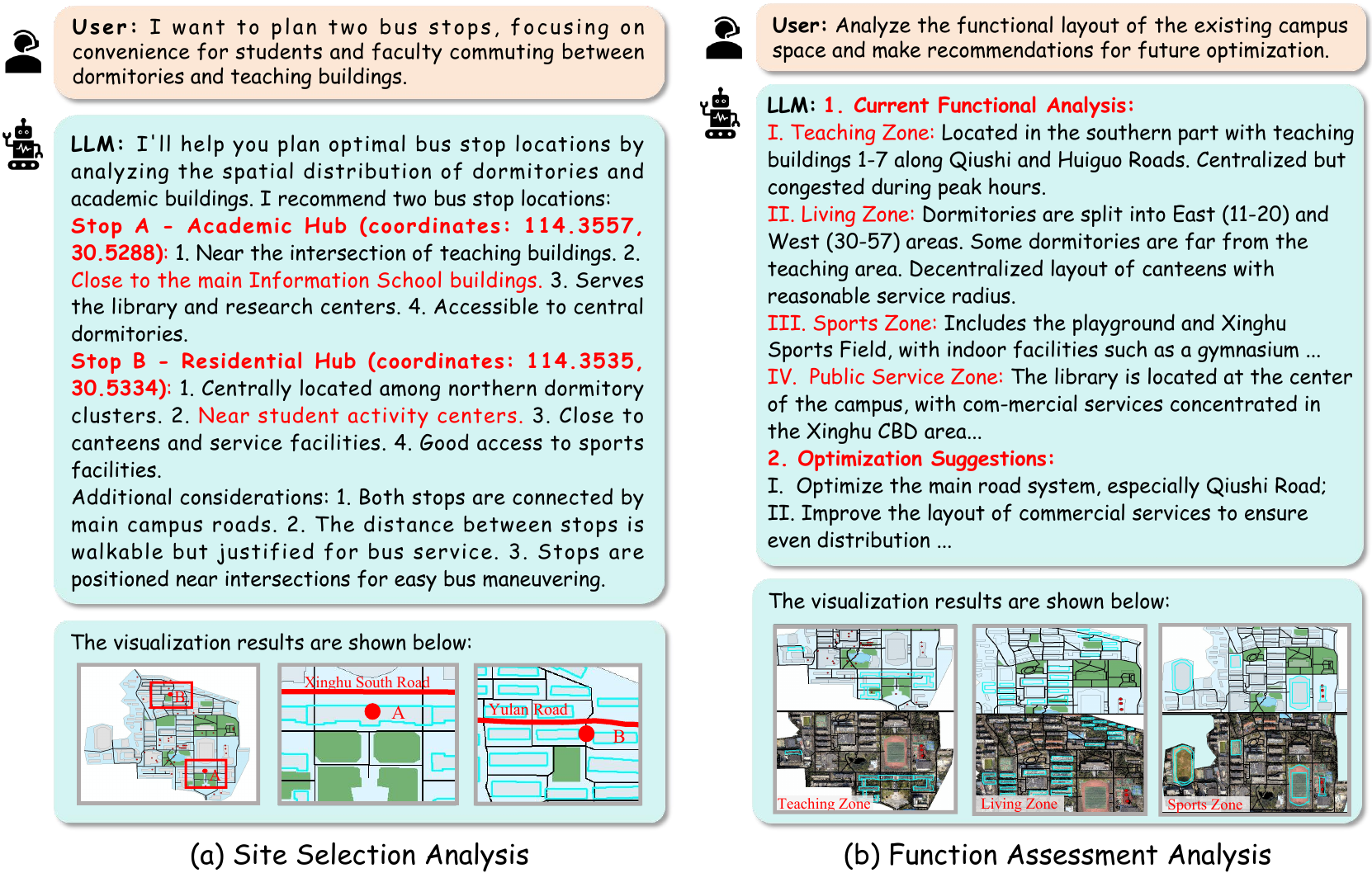}
    \includegraphics[width=0.92\linewidth]{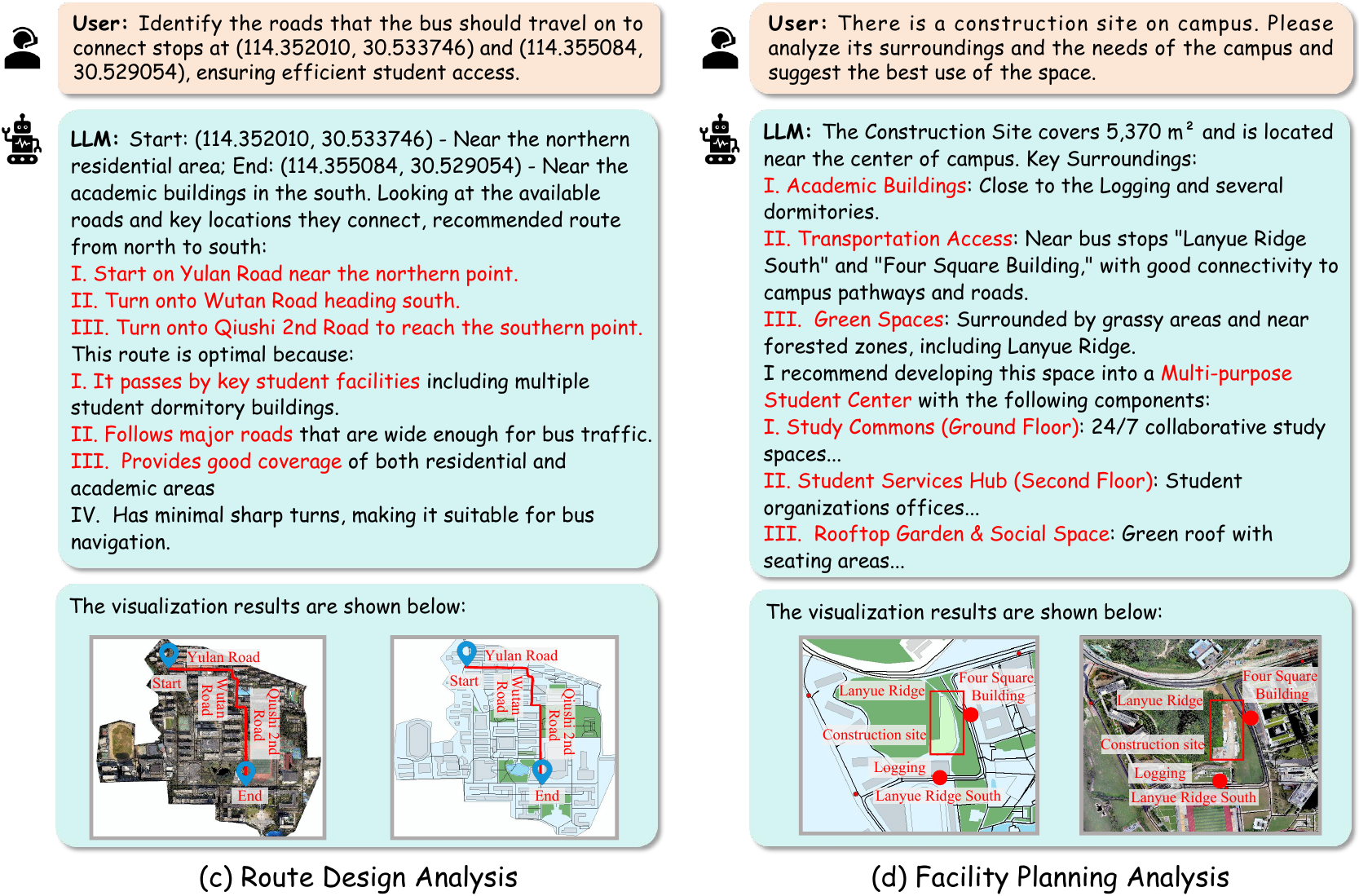}
    \caption{Application of SpatialLLM in urban planning and development.} 
    \label{fig:advance_12}
\end{figure*}

\begin{figure*}[!ht]
    \centering
    \includegraphics[width=0.92\linewidth]{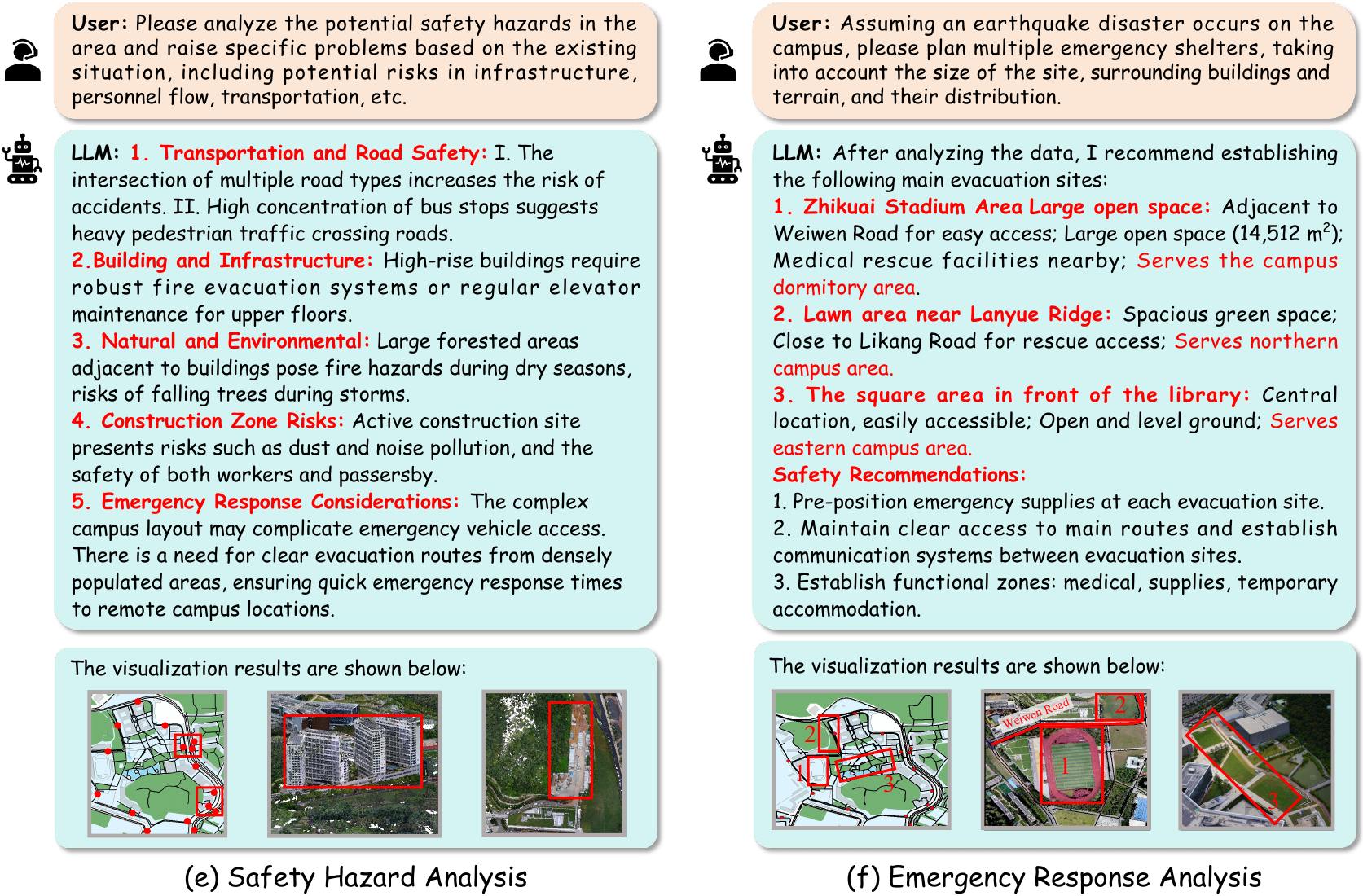} 
    \includegraphics[width=0.92\linewidth]{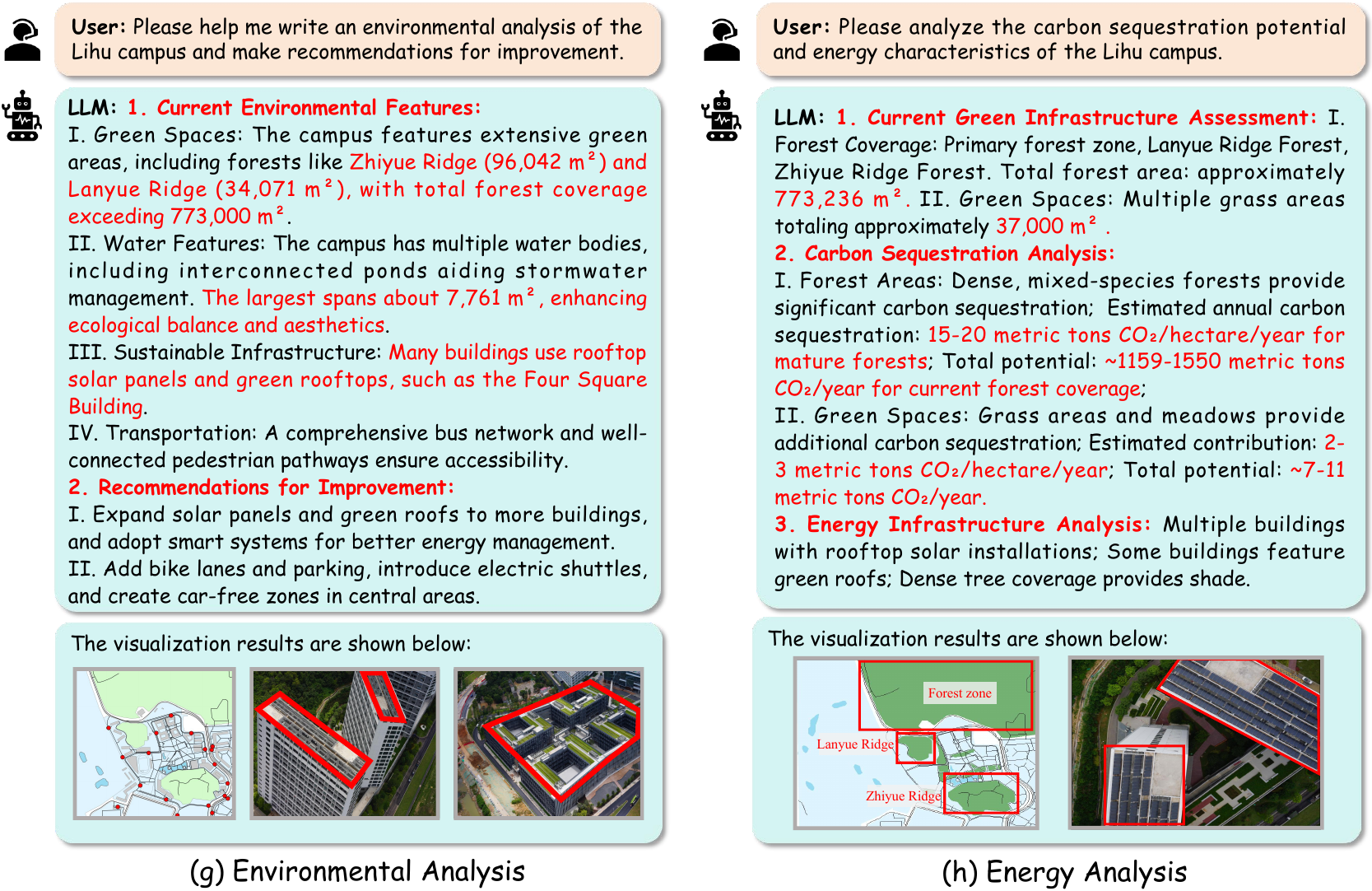} 
    \caption{Application of SpatialLLM in urban analysis of safety and ecological domains.}
    \label{fig:advance_34}
\end{figure*}

\begin{figure*}[!ht]
    \centering
    \includegraphics[width=0.92\linewidth]{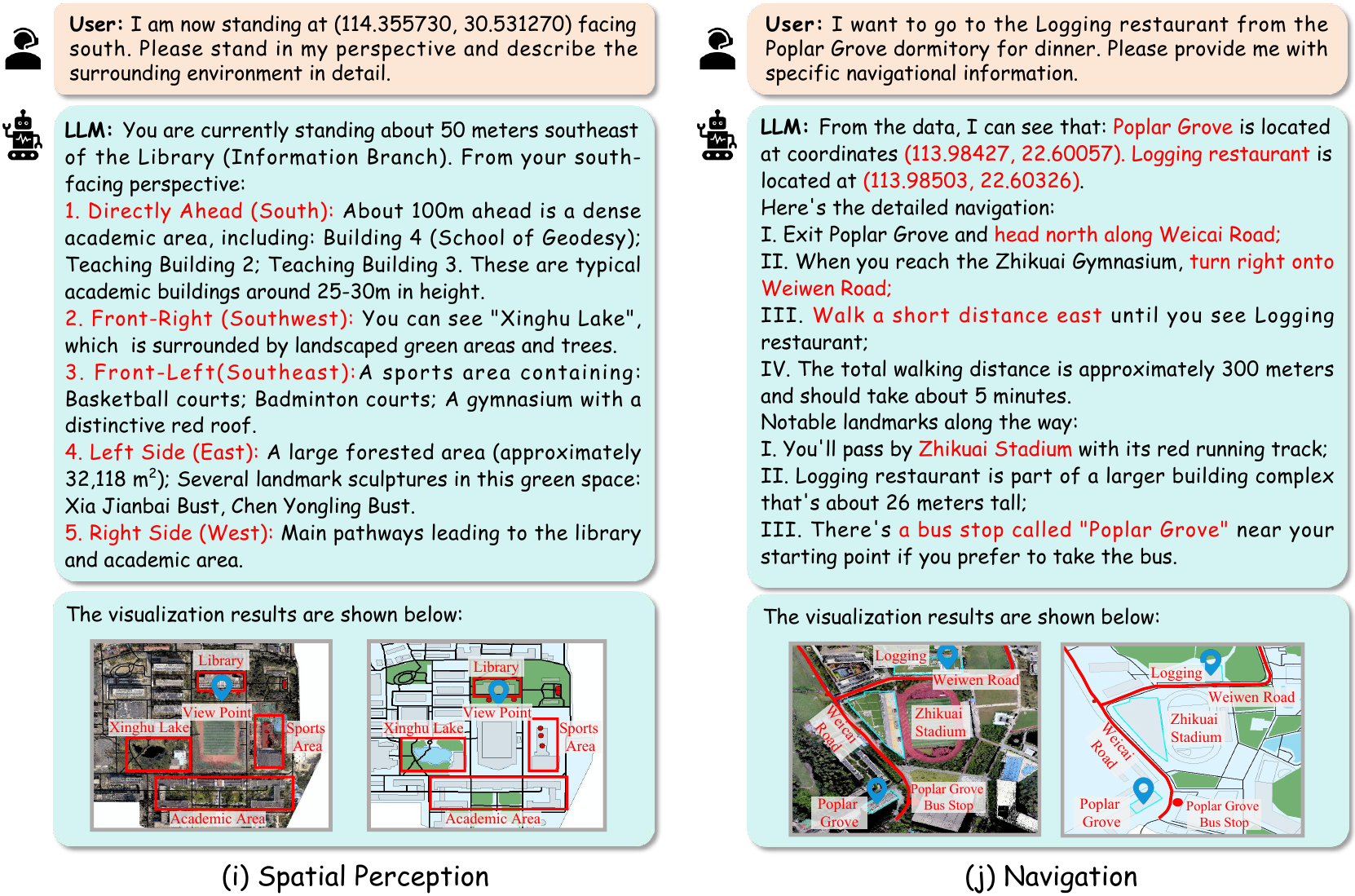}
    \includegraphics[width=0.92\linewidth]{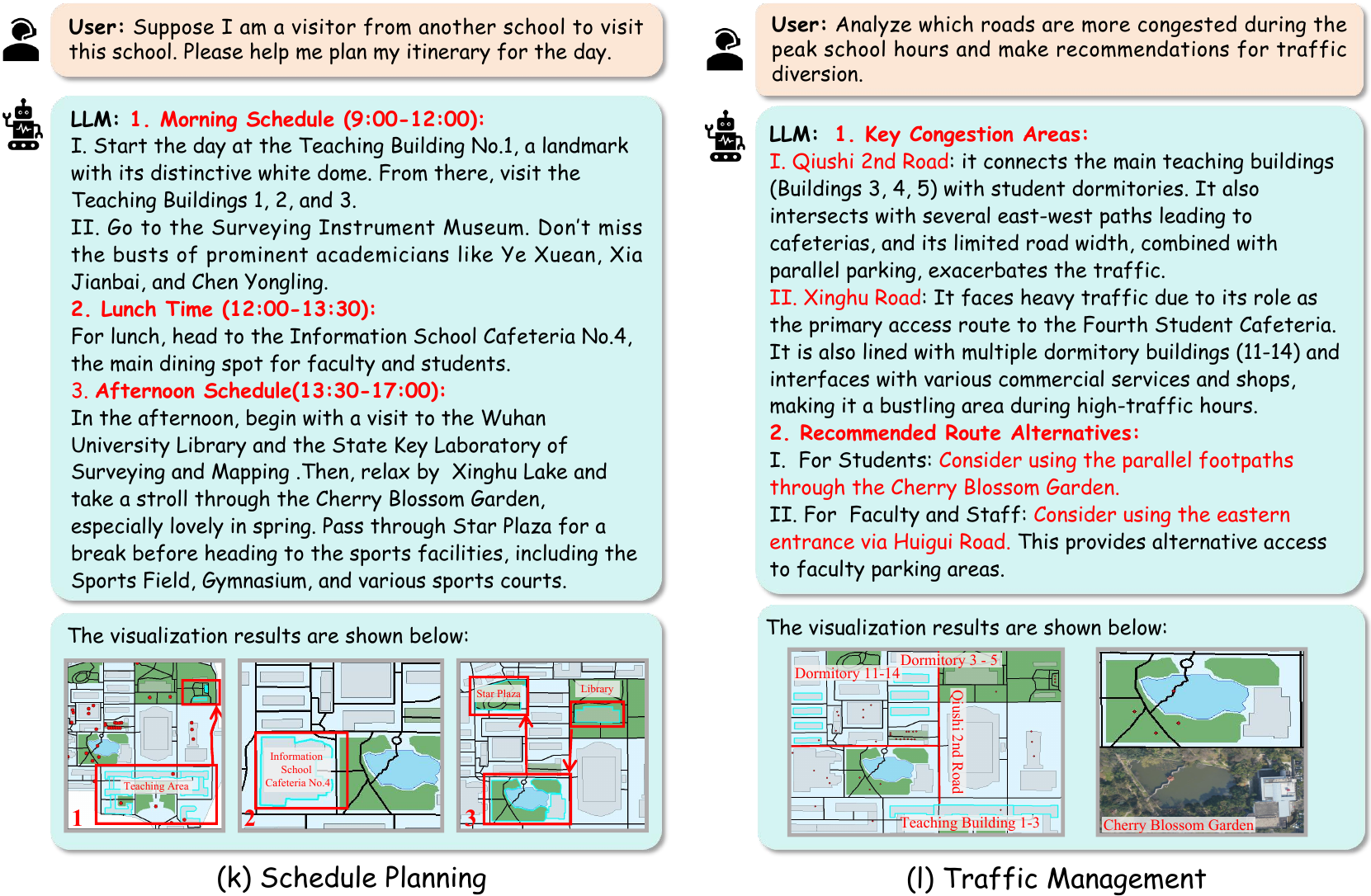}
    \caption{Application of SpatialLLM in traffic management domains.}
    \label{fig:advance_56}
\end{figure*}

Advanced spatial intelligence tasks extend beyond basic perception capabilities to require comprehensive analysis and complex reasoning about urban environments. Our evaluation reveals that SpatialLLM not only excels in fundamental scene perception tasks as shown in previous sections, but also effectively tackles advanced spatial intelligence applications across multiple domains. Figs.~\ref{fig:advance_12} to ~\ref{fig:advance_56} illustrate the performance of SpatialLLM on a variety of complex tasks across several representative domains, including urban planning, urban analysis, and traffic management.

Fig.~\ref{fig:advance_12} shows SpatialLLM's capabilities in urban planning and development. In (a) Site Selection Alaysis, SpatialLLM provided reasonable site selection solutions by identifying the distribution of dormitory areas and teaching areas, and considering the distance between stations and road accessibility. In (b) Function Assessment Analysis, SpatialLLM effectively analyzed the functional zones of campus, identified current spatial distribution patterns, and proposed targeted optimization suggestions. In (c) Route Design Analysis, SpatialLLM successfully designed an optimal bus route between two specified coordinates, considering key factors such as facility access, area coverage. In (d) Facility Planning Analysis, SpatialLLM thoroughly analyzed the construction site's location and surroundings, then provided comprehensive recommendations for developing a multi-purpose student center based on campus needs.

Fig.~\ref{fig:advance_34} demonstrates SpatialLLM's applications in urban analysis on ecological and safety domains. In (e) Safety Hazard Analysis, SpatialLLM comprehensively identified potential safety risks across multiple categories, including transportation, infrastructure, and construction zones, while providing relevant safety recommendations. In (f) Emergency Response Analysis, SpatialLLM identified suitable emergency evacuation sites based on strategic locations, accessibility, and space requirements, while providing comprehensive safety recommendations for site management. In (g) Environmental Analysis, LLM accurately analyzes conditions such as green spaces and water bodies on campus and makes recommendations for improvement. In (h) Energy Analysis, SpatialLLM evaluated the campus carbon sequestration capacity and energy features by assessing forest coverage, green spaces, and energy infrastructure, providing detailed quantitative data for environmental impact.

Fig.~\ref{fig:advance_56} illustrates SpatialLLM's usages in traffic management. In (i) Spatial Perception, SpatialLLM effectively provided comprehensive environmental descriptions from the specified location, offering detailed observations in cardinal directions with key landmark identifications. In (j) Navigation, SpatialLLM accurately pinpointed the locations of both starting and ending points, provided step-by-step navigation instructions, and included landmarks along the way. In (k) Schedule Planning, SpatialLLM created a well-structured day itinerary for a campus visitor, organizing activities across morning, lunch, and afternoon periods while highlighting key campus landmarks and facilities. In (l) Traffic Management, SpatialLLM effectively identified congestion-prone areas during peak hours, explained their causes, and provided practical alternative routes for traffic diversion.

Note that unlike traditional approaches that require specialized geospatial analysis tools and domain expertise, SpatialLLM can perform complex spatial reasoning tasks through a unified text-based framework without any training, fine-tuning, or expert intervention, which provides a novel perspective for urban analysis and management.


\section{Conclusion}
\textbf{Summary}. In this work, we introduced SpatialLLM, a novel framework for addressing the challenges of spatial intelligence processing in large-scale and complex scenes. 
The cornerstone of SpatialLLM lies in its ability to generate structured descriptions of various scenes from raw spatial data. These descriptions can verify and enhance the spatial perception capabilities of pre-trained LLMs. To our best knowledge, SpatialLLM is the first to validate the feasibility of zero-shot execution of advanced spatial intelligence tasks such as urban planning and urban analysis. We hope this can provide a novel paradigm for urban spatial intelligence processing.

\textbf{Limitations}. SpatialLLM converts complex outdoor scenes into SSD for LLMs to perform various tasks. However, as the scene range grows, the SSD length might exceed LLM's context limits, leading to decreased model performance and higher computational costs. Token compression strategies or Retrieval-Augmented Generation (RAG) techniques might be useful, which we leave for future work.

% \clearpage
%% The file named.bst is a bibliography style file for BibTeX 0.99c
\bibliographystyle{named} % named,unsrt
\bibliography{reference}
\clearpage
\section{Appendix}

\subsection{Prompt Design}
% Fig.~\ref{fig:caption prompt} shows the system prompts used for Self-caption and Surrounding-caption generation in Section.~\ref{sec:obj info}, which guide the vision-language model to produce detailed descriptions of target objects and their contexts. Fig.~\ref{fig:QA prompt} presents the system prompt for spatial perception QA tasks, which explicitly instructs the model on the structure of scene descriptions and defines a standardized output format to facilitate quantitative evaluation of spatial reasoning capabilities.
List.~\ref{lst:self-caption} and ~\ref{lst:surrounding-caption} shows the system prompts used for Self-caption and Surrounding-caption generation in Section.~\ref{sec:obj info}, which guide the vision-language model to produce detailed descriptions of target objects and their contexts. List.~\ref{lst:QA} presents the system prompt for spatial perception QA tasks, which explicitly instructs the model on the structure of scene descriptions and defines a standardized output format to facilitate quantitative evaluation of spatial reasoning capabilities.

\definecolor{codebg}{rgb}{0.96,0.96,0.96}  % 浅灰色
\lstset{
    numberstyle=\small  ,      % 行号字体大小
    breaklines=true,        % 允许自动换行
    captionpos=b,          % 标题位置在底部
    backgroundcolor=\color{codebg},  % 设置背景颜色
    basicstyle=\ttfamily\footnotesize  ,  % 基本字体样式
    breakindent=2pt,        % 换行后的缩进量
    lineskip=0.6pt,  
    keywordstyle=\color{blue},   % 关键词颜色
    commentstyle=\color{green},  % 注释颜色
    stringstyle=\color{red},     % 字符串颜色
    moredelim=[is][\color{red}]{@r}{@r}, 
    moredelim=[is][\color{blue}]{@b}{@b},       % 蓝色文本
    moredelim=[is][\itshape]{@it}{@it},         % 倾斜文本
    moredelim=[is][\bfseries]{@bf}{@bf},        % 加粗文本
    moredelim=[is][\bfseries\itshape]{@bfit}{@bfit},  % 加粗倾斜
    escapechar=| 
}
\begin{lstlisting}[caption = {The system prompt of Self-caption.},label={lst:self-caption}]
Self-caption prompt = "You are an advanced AI image analysis system capable of @rgenerating detailed captions for object highlighted by a red bounding box@r in images. The red bounding box highlights a @b{fclass}@b. Please generate a detailed yet concise caption describing the @b{fclass}@b, focusing on:
1. Describe its @roverall appearance@r, including color and shape
2. Highlight any @rvisible details or surface characteristics@r (e.g., patterns, textures, markings)
3. Focus on any unique features @rin the top or upper portion@r of the @b{fclass}@b
Please keep the description brief and to the point."
\end{lstlisting}

\begin{lstlisting}[caption = {The system prompt of Surrounding-caption.},label={lst:surrounding-caption}]
Surrounding-caption messages = "You are an advanced AI image analysis system capable of @rgenerating detailed captions for the surroundings of object highlighted by a red bounding box@r in images. The red bounding box highlights a @b{fclass}@b. Please generate a detailed yet concise caption describing the @b{fclass}@b, focusing on:
1. Describe the @rsurrounding objects@r (e.g., structures, vegetation and roads)
2. Describe the @rspatial relationship@r between the @b{fclass}@b and its surroundings
3. Note any @rsignificant features@r in the background and overall landscape
Please keep the description brief and to the point."
\end{lstlisting}

\begin{lstlisting}[caption = {The system prompt of spatial perception QA.},label={lst:QA}]
Low-level perception QA prompt = Based on the data provided by the user, you can @rperform many spatial reasoning tasks@r. In addition to some basic information, the structured text also includes several specialized fields:
- @rID@r: Unique identifier for each geographic object (e.g.,"1317798")
- @rbbox@r: The coordinates of the bottom-left and top-right corners of the minimum bounding rectangle of the polygon.
- @rVisual information@r: Detailed description of the object's physical appearance and immediate environment.
- @rSpatial Relationship@r: Direction and distance to surrounding objects
- @rGeographic Topology Relationship@r: Geographic information about the object's surroundings, including: Adjacent roads, Points of interest (POIs) with their distances.
Your answers must rely strictly on this data structure, and your output should follow this format: @rOption#Reasoning process.@r If none of the options are correct, output: @rF#Reasoning process@r
Example: User Question: If I am at building A, in which direction should I walk to reach building B?
A. Northwest  B. Southwest  C. Southeast  D. Northeast
Answer: C#Based on the data, ...., making option C the correct answer
\end{lstlisting}

\subsection{Structured Object Description}
We present a representative example of the structured object descriptions in Section~\ref{sec:SSD}, which structurally contains the multi-source object information and object relationship information.

\definecolor{key}{RGB}{20,20,20}
\definecolor{cont}{RGB}{150,150,150}
\definecolor{note}{RGB}{25,159,225}
\definecolor{rela}{RGB}{225,100,25}

\begin{algorithm}
\caption{Structured Object Description}
\label{alg:polygon-type}
\begin{algorithmic}[1]
\STATE {\color{key}\textbf{``104372384''}}:  {\color{note}  \ \ \# Object ID}
\STATE \{
\STATE \quad {\color{rela} \# Identity information from map data}
\STATE \quad {\color{key}\textit{``Name''}}: {\color{cont}``Building 1'',} {\color{note}  \ \ \# Object name}
\STATE \quad {\color{key}\textit{``Fclass''}}: {\color{cont}``building'',}{\color{note}   \ \ \# Object class}
\STATE \quad {\color{key}\textit{``Type''}}: {\color{cont}``dormitory'',}{\color{note}   \ \ \# Object type}
\STATE \quad {\color{rela} \# Geometric information from point clouds}
\STATE \quad {\color{key}\textit{``Center''}}: {\color{cont}[114.34, 30.51],} {\color{note}   \# Object center coordinates} 
\STATE \quad {\color{key}\textit{``Height''}}: {\color{cont}28 m,}{\color{note}   \ \ \# Object height}
\STATE \quad {\color{key}\textit{``Area''}}: {\color{cont}1200 m$^2$,} {\color{note}  \ \ \# Object area}
\STATE \quad {\color{key}\textit{``Volume''}}: {\color{cont} 33600 m$^3$,} {\color{note}  \ \ \# Object volume}
\STATE \quad {\color{note} \# Bounding box coordinates of object}
\STATE \quad {\color{key}\textit{``Bbox''}}: {\color{cont}[[114.33, 30.50], [114.35, 30.52]] }
\STATE \quad {\color{rela} \# Visual information from images}
\STATE \quad {\color{key}\textit{``Visual information''}}: {\color{cont}``A rectangular, multi-story building....'',}
\STATE \quad {\color{rela} \# Spatial relationship information}
\STATE \quad {\color{key}\textit{``Spatial Relationship''}}: {\color{cont}[\{}
\STATE \quad\quad  {\color{cont}``Name'':``Building 2'',}
\STATE \quad\quad  {\color{cont}``Direction'':``North'',}
\STATE \quad\quad  {\color{cont}``Distance'':34 m,}
\STATE \quad\quad {\color{cont}\},...],} 
\STATE \quad {\color{rela} \# Geographic topology relationship information}
\STATE \quad {\color{key}\textit{``Geographic Topology Relationship''}}: {\color{cont}\{}
\STATE \quad\quad  {\color{cont}``Point-type'':[(``Rm gym Bus stop'', 42 m),...],}
\STATE \quad\quad  {\color{cont}``Polyline-type'':[``Ren Min Road'',...]\}} 

\STATE \}
\end{algorithmic}
\end{algorithm}
\end{document}